\documentclass[12pt,twoside,a4paper]{article}

\usepackage[bt,fs,english]{packages/ethasl}
\usepackage{acronym} 
\usepackage{graphicx} 
\usepackage{setspace}

\usepackage[utf8]{inputenc}
\usepackage[OT1]{fontenc}

\usepackage{a4}

\usepackage{fancyhdr}

\usepackage{textcomp}\usepackage{gensymb}

\usepackage[english]{isodate}

\usepackage{multicol}

\usepackage{graphicx}

\usepackage{subfigure}

\usepackage[numbers]{natbib}

\usepackage{booktabs}
\usepackage{array}
\usepackage{multirow}

\usepackage{color}
\usepackage{colortbl}
\definecolor{black}{rgb}{0,0,0}
\definecolor{white}{rgb}{1,1,1}
\definecolor{darkred}{rgb}{0.5,0,0}
\definecolor{darkgreen}{rgb}{0,0.5,0}
\definecolor{darkblue}{rgb}{0,0,0.5}

\usepackage{amsmath}
\usepackage{amssymb}

\usepackage{nicefrac}

\usepackage{upgreek}

\usepackage{isomath}

\usepackage{units}

\usepackage{rotating}

\usepackage{pdfpages}
\includepdfset{pages={-}, frame=true, pagecommand={\thispagestyle{fancy}}}

\usepackage{pgfgantt}

\PassOptionsToPackage{hyphens}{url}\usepackage{hyperref}

\usepackage{cleveref}

\begin{document}
\title{Cone Detection}
\subtitle{Using a Combination of LiDAR and Vision-Based Machine Learning}

\studentA{Nico Messikommer}
\studentB{Simon Schaefer}

\supervisionA{Renaud Dub\'e}
\supervisionB{Mark Pfeiffer}

\projectYear{\the\year} 

\maketitle
\pagestyle{plain}
\pagenumbering{roman}

\newpage

\onehalfspacing
\section*{Abstract}
\addcontentsline{toc}{section}{Abstract}

The classification and the position estimation of objects become more and more relevant as the field of robotics is expanding in diverse areas of society. In this Bachelor Thesis, we developed a cone detection algorithm for an autonomous car using a LiDAR sensor and a colour camera. By evaluating simple constraints, the LiDAR detection algorithm preselects cone candidates in the 3 dimensional space. The candidates are projected into the image plane of the colour camera and an image candidate is cropped out. A convolutional neural networks classifies the image candidates as cone or not a cone. With the fusion of the precise position estimation of the LiDAR sensor and the high classification accuracy of a neural network, a reliable cone detection algorithm was implemented. Furthermore,  a path planning algorithm generates a path around the detected cones. The final system detects cones even at higher velocity and has the potential to drive fully autonomous around the cones.
\newpage
\section*{Preface}
\addcontentsline{toc}{section}{Preface}

We would like to thank a number of people who have encouraged and helped us in writing this Bachelor Thesis. We are very proud of the achieved results and we appreciate the support we received from all sides.
\vspace{0.5cm}
\newline
We are much obliged to Prof. Dr. Roland Y. Siegwart for his support, for the confidence in our project and for providing the opportunity to execute it.
\vspace{0.5cm}
\newline
Special thanks to our supervisors and PhD candidates Renaud Dub\'e and Mark Pfeiffer. Their broad knowledge and experiences in perception, machine learning, control and path planning as well as their very professional and friendly way of support had a wide influence on the results and also on our education. They persistently helped us with many complex questions and motivated us to extend the scope of the project into the right directions.
\vspace{0.5cm}
\newline
We are very thankful to all members of the team ARC and are happy to have spent a year together working as a team on our shared vision of Project ARC. We have formed a very capable team and helped each other to realize our ideas.
\vspace{0.5cm}
\newline
Most importantly, we are very grateful to all our friends and family members who supported us during the whole project and helped us to achieve our aims.
\vspace{0.5cm}
\newline
Zürich in June 2017,
\newline
Nico Messikommer and Simon Schaefer

\newpage

\setcounter{tocdepth}{2}

\tableofcontents
\cleardoublepage


\section*{Acronyms and Abbreviations}
\addcontentsline{toc}{section}{Acronyms and Abbreviations}

\begin{acronym}[ORB-SLAM2]
\setlength{\itemsep}{-\parsep}
	\setlength{\itemsep}{-0.2cm}
	\acro{ARC}{Autonomous Racing Car}
	\acro{CNN}{Convolutional Neural Net}
	\acro{CPU}{Central Processing Unit}
	\acro{ETH}{Swiss Federal Institute of Technology}
	\acro{GNSS}{Global Navigation Satellite System}
	\acro{GUI}{Graphical User Interface}
	\acro{HSV}{Hue, Saturation, Value Colorspace}
	\acro{LAB}{Lightness, Green-Red and Blue-Yellow Colourspace}
	\acro{LiDAR}{Light Detection And Ranging}
	\acro{LOGITECH}{Logitech Webcam c920}
	\acro{MPC}{Model Predictice Control}
	\acro{ORB-SLAM2}{Oriented FAST and Rotated BRIEF - Simultaneous Localisation And Mapping 2}
	\acro{RGB}{Red, Green, Blue Colorspace}
	\acro{ROC}{Receiver-Operating-Characteristic}
	\acro{ROS}{Robot Operating System}
	\acro{ROVIO}{Robust Visual Inertial Odometry}
	\acro{SLAM}{Simultaneous Localisation And Mapping}
	\acro{SVM}{State Vector Machine}
	\acro{VI}{Visual Inertial}

\end{acronym}

\newpage
\mbox{}
\newpage

\cleardoublepage

\pagestyle{fancy}
\pagenumbering{arabic}

\section{Introduction}

\subsection{Vision and Motivation}
Detecting objects reliably is a crucial and necessary part of fully autonomous driving, especially for anticipating the actions of other road users, e.g. cyclists and pedestrians. While current object detection approaches have to deal with issues regarding accuracy or computational power, we want to solve these problems with combining sensors, exemplary demonstrated on the basis of cone detection \footnote{As demonstrated in \ref{sec:PerformanceAnalysis} we developed a method providing high accuracy and recall with low computational demand compared to merely camera or \ac{LiDAR} based approaches.}.
\newline
\newline
The motivation behind our Bachelor Thesis arose during our participation in the project \ac{ARC}. The goal of Project \ac{ARC} was to develop a vehicle capable of driving autonomously on a Swiss mountain pass using a vision-based Teach \& Repeat method \cite{furgale}. To autonomously drive a given path, this path first needs to be manually driven and teached to the vehicle. This methode can be useful for repeating frequent routes such as commuting to work and public transportation. This approach however requires a-priori knowledge of the environment which is not always practical in real-world scenarios.
\newline
Within the scope of our Bachelor Thesis, we want to take a further step towards fully autonomous driving, i.e. to develop a system independent of the Teach \& Repeat method and which does not rely on prior knowledge about the environment. To this end, we decided to focus on a fully autonomously slalom drive around cones. In order to drive a slalom, the car has to reliably detect the cones marking the course. There also lays our main focus; the development of a cone detection using a comination of a colour camera and a \ac{LiDAR} sensor by applying simple constraints for the preselection of candidates and a machine learning algorithm to identify the cones.

\subsection{Goals}
As our main focus lays on the detection of cones, we want to develop a system able to identify cones placed on a road with a low false positive rate and a high accuracy in the estimated cone position. A low false positive rate is crucial due to the fact, that a false detection can lead to an eroneous decision by the vehicle. The cone detection should even provide correct results in a busy environment, where plenty of objects are located beside the cones. As an additional goal, the detected cones should be stored in a global 2D cone gridmap in order to plan a global path around the cones. To compute the global position of the cones, the state of the vehicle first needs to be estimated in real time. At the beginning of our Bachelor Thesis, we decided to employ the same state estimation technique which was used for the teach part during Project \ac{ARC} as we assumed this state estimation to be accurate enough for driving around cones. Furthermore, the second additional goal was to generate a path around the detected cones. In the end, the system should be capable of driving fully autonomously around cones.

\subsection{Platform}

\begin{figure}[h]
   \centering 
   \includegraphics[width=0.7\textwidth]{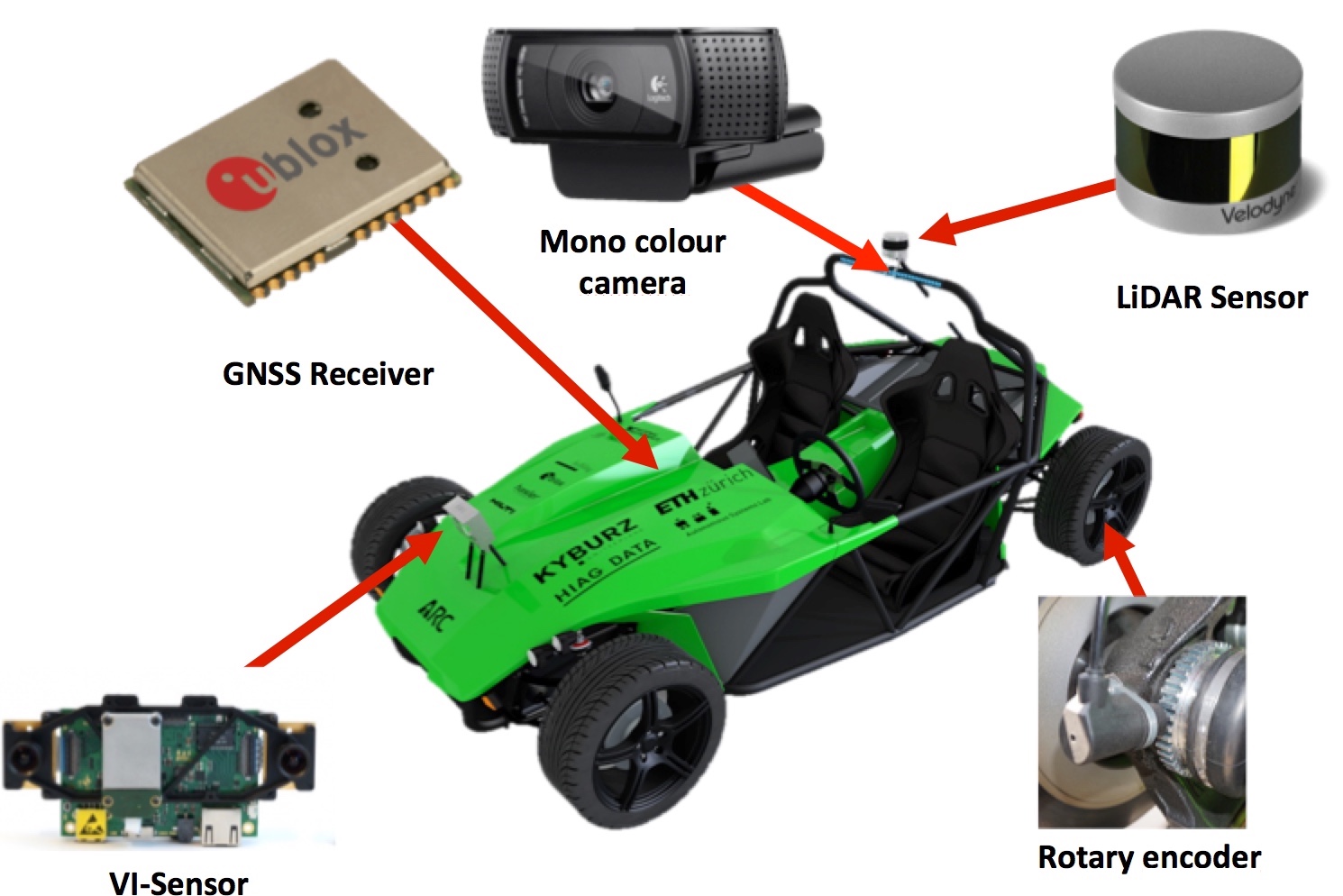} 
   \caption{eRod vehicle including all sensors}
   \label{fig:eRod}
\end{figure}

As this Bachelor thesis is conducted within the framework of Project \ac{ARC}, our developed system is implemented on an electric car, an eRod designed and manufactured by Kyburz Switzerland AG, used during Project \ac{ARC}. The modified eRod has an electrically actuated steering and braking system. Due to the car's application in Project \ac{ARC}, we were restricted in positioning the sensors, shown in Figure \ref{fig:eRod}. As further explained in \ref{sec:AlternativeApproaches}, we chose a combination of a \ac{LiDAR} sensor and \ac{LOGITECH} in order to detect cones, while for the state estimation merely rotary and steering angle encoders are used.

The 3D \ac{LiDAR} sensor (Velodyne VLP-16) emits rays of wavelength 903 nm with 5-20 Hz in every horizontal direction and between $\pm$ 15\degree \, in vertical direction, with an angle resolution of 2\degree \, in vertical and 0.4\degree \, in horizontal direction. Surfaces in the stated range reflect the laser beam. Using the measured time difference between the outgoing and incoming laser beam, the horizontal and the pitch angle, a point cloud can be constructed. In addition, the \ac{LiDAR} sensor provides information about the intensity of the reflected beams. The intensity varies according to the reflectivity of the reflecting surface.
\newline
The colour camera provides \ac{RGB} mono images in 1080p resolution recorded with a rate of 30 Hz. As normally used for video calling applications, the camera adjusts focus as well as exposure automatically. To ensure a sharp image of the cones in front of the car, the focus was set to infinity. Since the colour camera is placed directly below the \ac{LiDAR} sensor, a transformation from world to image plane using a pinhole camera model requires only a translation.

\subsection{State of the Art}

\begin{figure}[h]
   \centering 
   \includegraphics[width=0.9\textwidth]{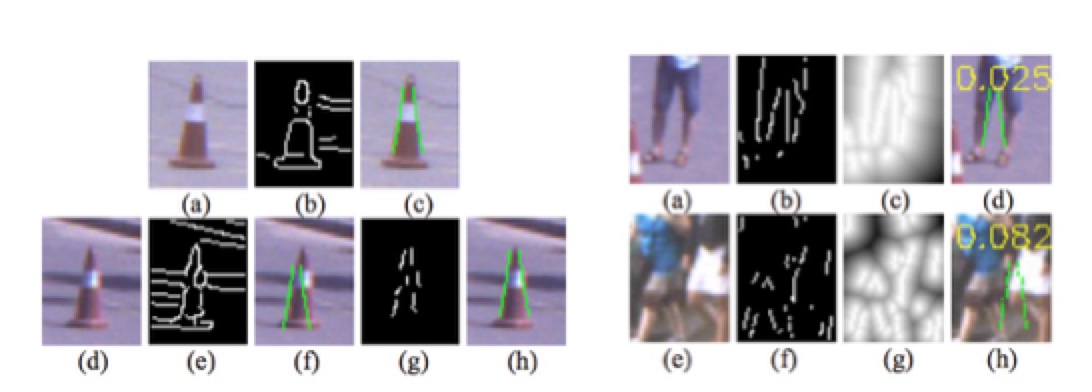} 
   \caption{Cones detection using chamfers according to \cite{cone_line_detection}}
   \label{fig:Hoangzhou_Detection}
\end{figure}

To the best of our knowledge, there is little scientific work which was published on the topic of real-time cone detection for autonomous driving. In 2004, F. Lindner, U. Kressel and S.Kaelbarer \cite{trafficSign} presented merely vision based approach using single-pixel-classifier, i.e. a detection of specific colours and shapes, for a real-time traffic sign detector. In \cite{trafficSign} an object classification was already seperated from detection, so that the classifier has to evaluate solely preselected images. As an accurate position estimation is not necessary for traffic sign detecting, this merely vision based approach provided a high accuracy in detection but was imprecise in estimating the objects position. For improve the accuracy of position estimation, H. Yong and X. Jianru \cite{cone_line_detection} propose an approach based on the fusion of radar and camera. In this approach, the cones are treated as triangles and the triangle's chamfers are detected using chamfer matching as shown in Figure \ref{fig:Hoangzhou_Detection}.

Since this approach allows to detect cones at different scales (i.e. distances), it has a recall rate of 90 \%, but as shown in Figure \ref{fig:Hoangzhou_Detection}, this method is vulnerable to shapes enclosing similar chamfers. Hence, a low accuracy, 69.2 \%, is achieved. 

\subsection{Structure of the Report}
In section "Alternative Approaches" \ref{sec:AlternativeApproaches}, the advantages and disadvantages of only vision based and only \ac{LiDAR} based methods for cone detection are elaborated. Afterwards, in chapter "System" \ref{sec:System} our components are introduced one by one. Following, the chapter "Experimental Results" \ref{sec:PerformanceAnalysis} contains the performance analysis of the cone detection as well as of the overall system driving around cones. In section "Conclusion" \ref{sec:Conclusion}, a short summary and an outlook is presented.

\newpage
\section{Alternative Approaches}
\label{sec:AlternativeApproaches}

The sensor configuration provided by Project \ac{ARC} includes three exteroceptive sensors for environment perception, the \ac{LiDAR} sensor, the \ac{VI}-sensor and the colour camera. As the \ac{VI} sensor records black-and-white, the rich colour information is lost. Furthermore, due the inclination of the \ac{VI}-sensor towards the road, a small section far away from the camera is captured. In order to detect cones, we decided therefore to evaluate three possible configuration: LiDAR only, vision only and a combination of both. Due to the goal of fully autonomously driving a path through cones without further knowledge of the environment, a reliable, real-time and computational efficient classification algorithm is necessary. Hence, in the following the advantages and disadvantages of each approach are elaborated.

\subsection{Only LiDAR based Cone Detection}

\begin{figure}[h]
   \centering 
   \includegraphics[width=0.42\textwidth]{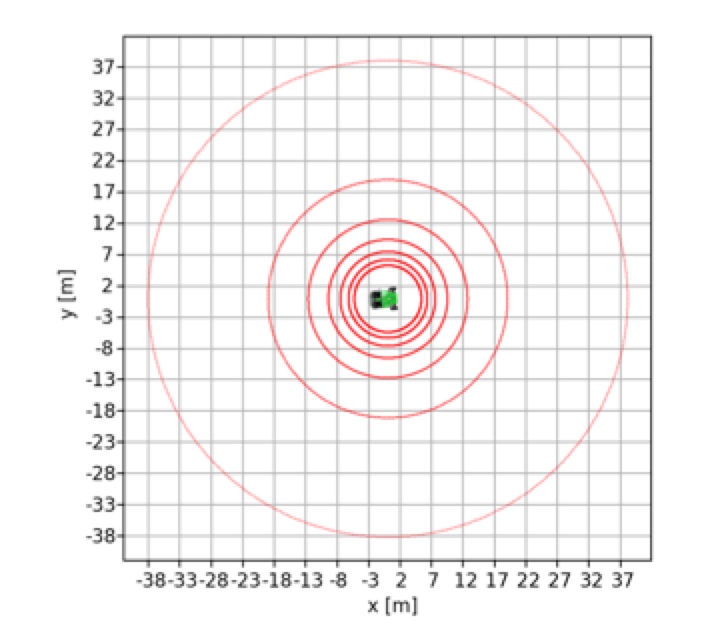} 
   \caption{Top view of the spatially dispersed \ac{LiDAR} readings}
   \label{fig:Lidar_Resolution}
\end{figure}

Due to its position at the top of the car and its limited vertical and horizontal angular resolution, the main challenge in a \ac{LiDAR}-based cone detection implemented on our system certainly is handling the sparseness of the \ac{LiDAR} sensor measurements. Figure \ref{fig:Lidar_Resolution} illustrates this scenario with a top-down view on the LiDAR reading pattern. Therefore, a cone can potentially lay between two reading lines invisile for a limited time to the laser. Assuming the shape of a cone as a pyramid, a cone is hit by less than 5 laser points from a distance more than 5.84 m and less than three laser points from a distance more than 7.33 m. 

\begin{figure}[h]
   \centering 
   \includegraphics[width=0.75\textwidth]{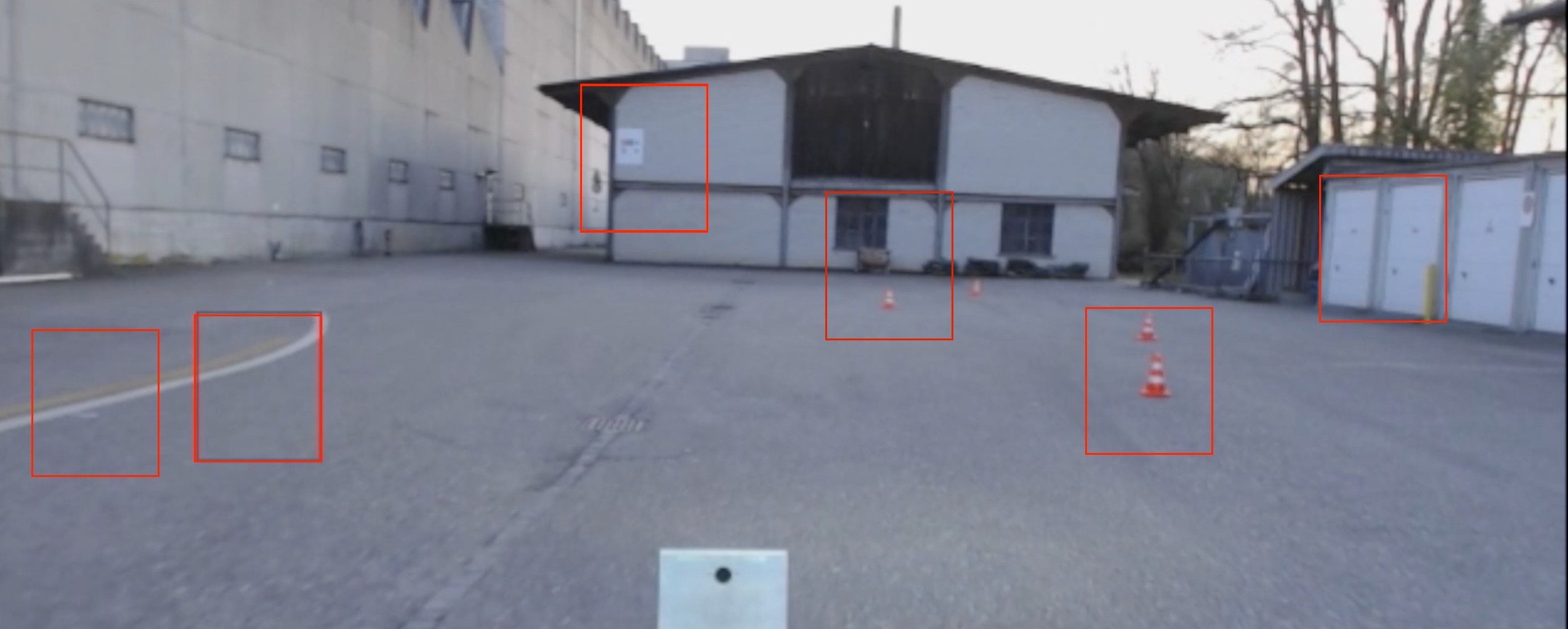} 
   \caption{False positives using intensity selection}
   \label{fig:Lidar_False_Positives}
\end{figure}

Having the goal to detect the cone without ambiguity and as far away from the car as possible, neither a spatial segmentation as in \cite{lidar_segmentation} nor a definite object classification using a \ac{SVM} approach proposed by M. Himmelsbach \cite{lidar_classification} is feasible. Due to sparsity, an another approach is the definition of simple constraints, which will be further described in section \ref{sec:LiDARDetection}, for the \ac{LiDAR} sensor points, so that cones can be detected up to a distance of 38 m. Because of the highly reflective surface, the intensity values can convey valuable information for identifying cones. But since a high reflectivity is not a property unique for cones, an high false positive rate is the result as shown in Figure \ref{fig:Lidar_False_Positives}.

Hence, an only \ac{LiDAR}-based approach is not sufficient for cone detection. Although by defining simple constraints for \ac{LiDAR} sensor points, a high recall is achieved. Therefore, in our final system presented in Chapter \ref{sec:System} we adopt this approach for extracting cone candidates.

\subsection{Only Vision Based Cone Detection}

A vision-based approach using a colour camera provides more information for classification, such as details about texture or shape but basically no depth infromation. Furthermore, since the information is not discretised into spatially distributed points, sparseness is not an issue. As most vision-based approaches work well in binary classification (i.e. to decide whether an image contains a cone or not), a sliding window algorithm cropping images out of the overall image can be employed in order to provide classification candidates. Here, a trade-off between limiting the number of images evaluated per frame, and the accuracy in position estimation, determined both by the windows size, has to be found. The high computational cost is one of the biggest disadvantages of a sliding window approach. As we made preliminary experiments using this approach, the frame rate of the the colour camera had to be dropped to 2 Hz in order to obtain the same classification accuracy as our final cone detection. The usage of a stereo camera would improve the accuracy of the position estimation of the cones but has several disadvantages. The biggest disadvantages are the necessity to detect and identify a cone in two images as the same cone and the computational demanding implementation. A vision-based approach certainly is computational demanding when the same accuracy in position estimation of the cone should be achieved as a \ac{LiDAR}-based approach. Therefore, a computational efficient algorithm has to be developed which is precise in the detection of cones.
\newline
Next to the highly reflective surface, the cones we considered in our experiments exhibit an orange color as well as a triangular shape, both properties easily distinguishable from other objects. Hence, reasonable approaches for vision-based cone detection is to regard the cone as a triangle and calculate the enclosed angle or to search for objects with an orange color. In the following, the different approaches are evaluated based on prior conducted experiments.

\subsubsection*{Triangle Detection}

In order to detect the specific triangle shape of cones with the edges in the image obtained by applying a Canny Edge Detector, the \ac{RGB} image from the colour camera has to be converted into a grayscale image. Inputting the resulting edges in a Hough transformation locates and parameterises lines in the image. The specifc triangle shape of cones in the image can be found by computing and searching for the correct angle enclosed by two lines. Once a specific triangle shape of cones is detected, the image is classified as a cone. Altough this approach is very computational intensive, it works well for cones close to the camera. But as already shown by \cite{cone_line_detection}, a lot of other objects enclose a similar triangle resulting in a high false positive rate. 

\begin{figure}[h]
   \centering 
   \includegraphics[width=0.8\textwidth]{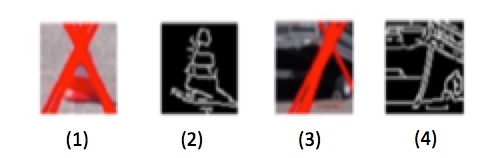} 
   \caption{ (1) correctly and (2) falsly classified image using triangle detection. (2) and (3) the detected lines.}
   \label{fig:Cone_Line_Detection_Examples}
\end{figure}

\subsubsection*{Colour Detection}

Another possible approach is to detect the colour of the cone. As a first step for colour detection, the \ac{RGB} image recorded by the colour camera is converted into the \ac{HSV} to separate colour from brightness information. Afterwards, the overall intensity of the orange color is calculated and compared to a threshold. A detection based on an object's colour is computationally highly efficient, but vulnerable since a lot of structures in the environment have a similar colour, as shown below.

\begin{figure}[h]
   \centering 
   \includegraphics[width=0.8\textwidth]{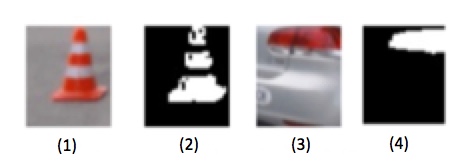} 
   \caption{(1) correcctly and falsly (3) classified image using colour detection. (2) and (4) the detected masks}
   \label{fig:Cone_Line_Detection_Examples}
\end{figure}

\subsubsection*{Combination of Line and Colour Detection}

\begin{figure}[h]
   \centering 
   \includegraphics[width=0.7\textwidth]{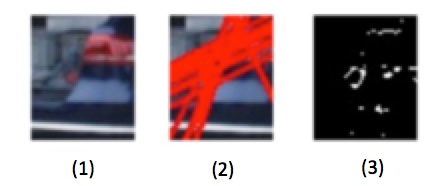} 
   \caption{Falsy classified backlight using combination of triangle and colour detection.}
   \label{fig:Cone_Line_Colour_Detection_Example}
\end{figure}

Combining triangle detection with colour detection improves the classification accuracy. However, as pictured in Figure \ref{fig:Cone_Line_Colour_Detection_Example}, this method can potentially wrongly detect challenging objects such as backlights. Nevertheless, since every cone exhibits these properties, a recall of about 100 \% is achieved. Hence, a combination of colour and simplified triangle detection works very well in preselecting images that could contain a cone.

\subsubsection*{Machine Learning Approaches}

For a more accurate classification, a more advanced method is required. Two frequently used methods in object classification are \ac{CNN}s as well as \ac{SVM}s. \ac{SVM}s weigh features in the input image in order to classify an image. In opposite to the previous approaches, a lot of training data is required but in return, a more distinct detection results. As shown in \cite{conv_net_vs_svm} with the example of pedestrian detection, the used \ac{CNN}s in the experiments are less computational demanding and have a smaller false positives rate than used \ac{SVM}s with the same overall accuracy. Thats why we decided to evaluate a \ac{CNN} and a fully connect neural network for the implementation on our system.

\newpage
\section{System}
\label{sec:System}

In order to fully autonomously drive a path between cones, we first need to  detect the cones, then estimate their position relative to the car, plan an optimal path around the cones and control the eRod according to the planned path. In this chapter, our approach will be presented.

\subsection{System Overview}
As our Bachelor Thesis is based on Project \ac{ARC}, we were able to adopt the state estimation as well as the high- and low-level control and focused on the development of a reliable cone detection.

\begin{figure}[h]
   \centering 
   \includegraphics[width=1.0\textwidth]{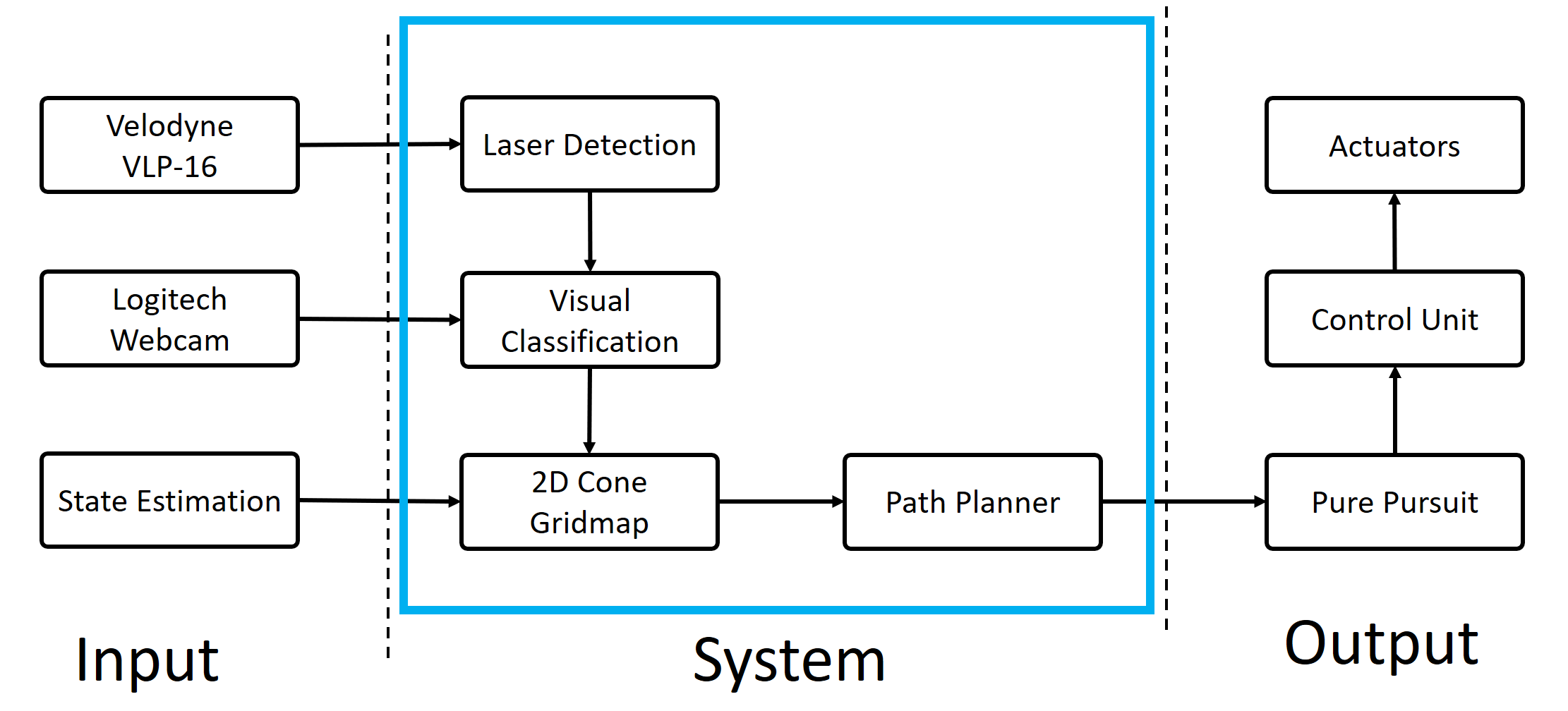} 
   \caption{System Overview}
   \label{fig:System_Overview}
\end{figure}

\begin{figure}[h]
   \centering 
   \includegraphics[width=0.75\textwidth]{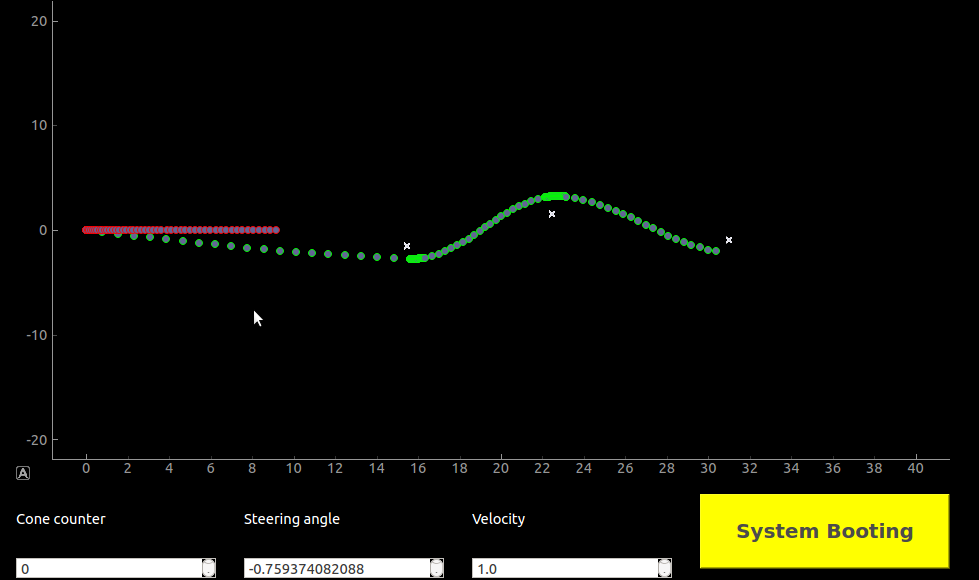} 
   \caption{\ac{GUI}: Crosses are detected cones, red and green dots belong to driven and planned path, respectivly.}
   \label{fig:Gui}
\end{figure}

In the final system, merely the colour camera, the \ac{LiDAR} sensor as well as rotary and steering encoder are used. As it can be seen in Figure \ref{fig:System_Overview}, for cone detection, we decided to combine the \ac{LiDAR} sensor with the colour camera in order to ensure both, a fast and accurate position estimation of a cone as well as a reliable classification. Hence, in our final system, we preselect cone candidates from all objects in the environment and estimate their global positions using the \ac{LiDAR} sensor and the pose of the car. An image of the candidates is then cropped out of the overall image by transforming the estimated object's position into the image plane using a pinhole camera model. Then, these cropped images are evaluated with a visual detection, if they contain a cone or not. After the visual evaluation, the cones are stored in a 2D cone gridmap, so that a path around the cones can be planned. In opposite to a visual-based sliding-window approach, only a few candidate images have to be evaluated. Hence, our system is capable of operating in real time even if an accurate, and therefore computational demanding, classifying algorithm is used. Nevertheless, in contrast to a merely \ac{LiDAR}-based approach (e.g. a spatial segmentation), the system is able to detect cones far away from the car, due to the fact that only one laser point reflected from a cone is needed for detection. Furthermore, to control the car while driving autonomously,  a \ac{GUI} was developed. Next to a button to change between autonomous and manual mode, a map shows all of the detected cones as well as the planned and driven path, as shown in Figure \ref{fig:Gui}.

\subsection{LiDAR Based Candidate Extraction}
\label{sec:LiDARDetection}
\begin{figure}[h]
   \centering 
   \includegraphics[width=0.8\textwidth]{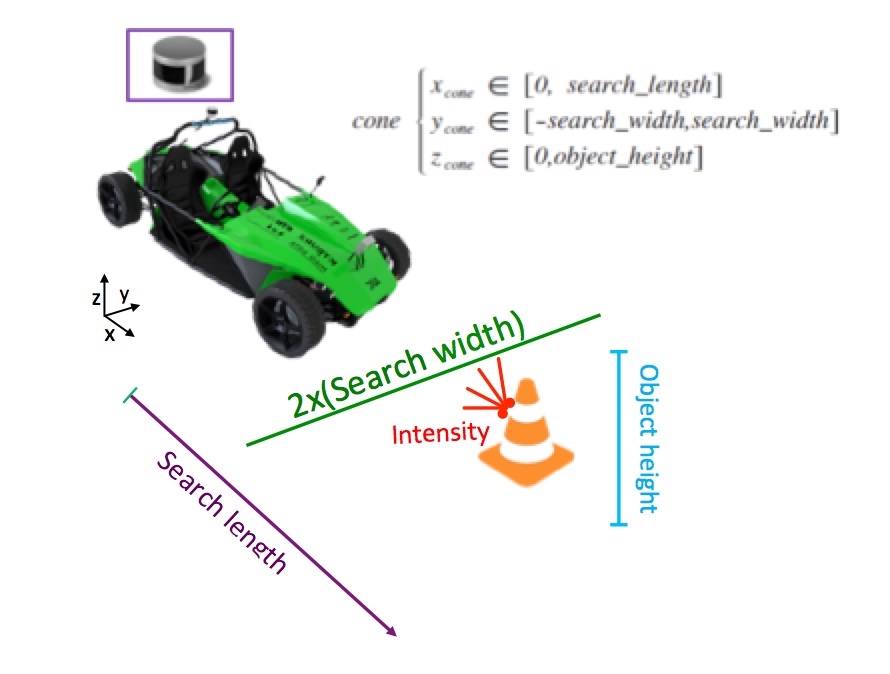} 
   \caption{Illustration of constraints for the \ac{LiDAR}-based candidate extraction}
   \label{fig:Lidar_Detection}
\end{figure}

The \ac{LiDAR} sensor captures a 3D pointcloud of the environment. Every point of the 3D pointcloud comprises to a local coordinate as well as to an intensity. In order to extract cone candidates out of all of these points as far away and as computational efficient as possible, simple constraints for the points had to be found. Due to the highly reflecting surface of a cone, we decided to use an intensity threshold as the most important criterion for the detection of points reflected by cones. As the \ac{LiDAR} detection has to handle both, the highly reflective (white-coloured) as well as the less reflective surface (orange-coloured) of the cone, the intensity threshold was set below the intensity of the points reflected by the orange surface resulting in filter detecting other reflective objects (e.g. number plates, streetlamps, backlights) as cones. That is why, in addition to an intensity threshold other geometric properties have to be fulfilled by a candidate, assuming the cones on the ground in front of the car, as shown in \ref{fig:Lidar_Detection}. 

With this approach, a candidate selection ensuring to detect every cone as soon as possible is implemented. The simple \ac{LiDAR} detection minimises the number of images required to be classified by the visual detection. As the visual classification obtains only images preselected by the \ac{LiDAR} detection, an high recall is much more important than precision for the \ac{LiDAR} detection.

\subsection{Image Based Cone Detection}
\begin{figure}
   \centering 
   \includegraphics[width=0.95\textwidth]{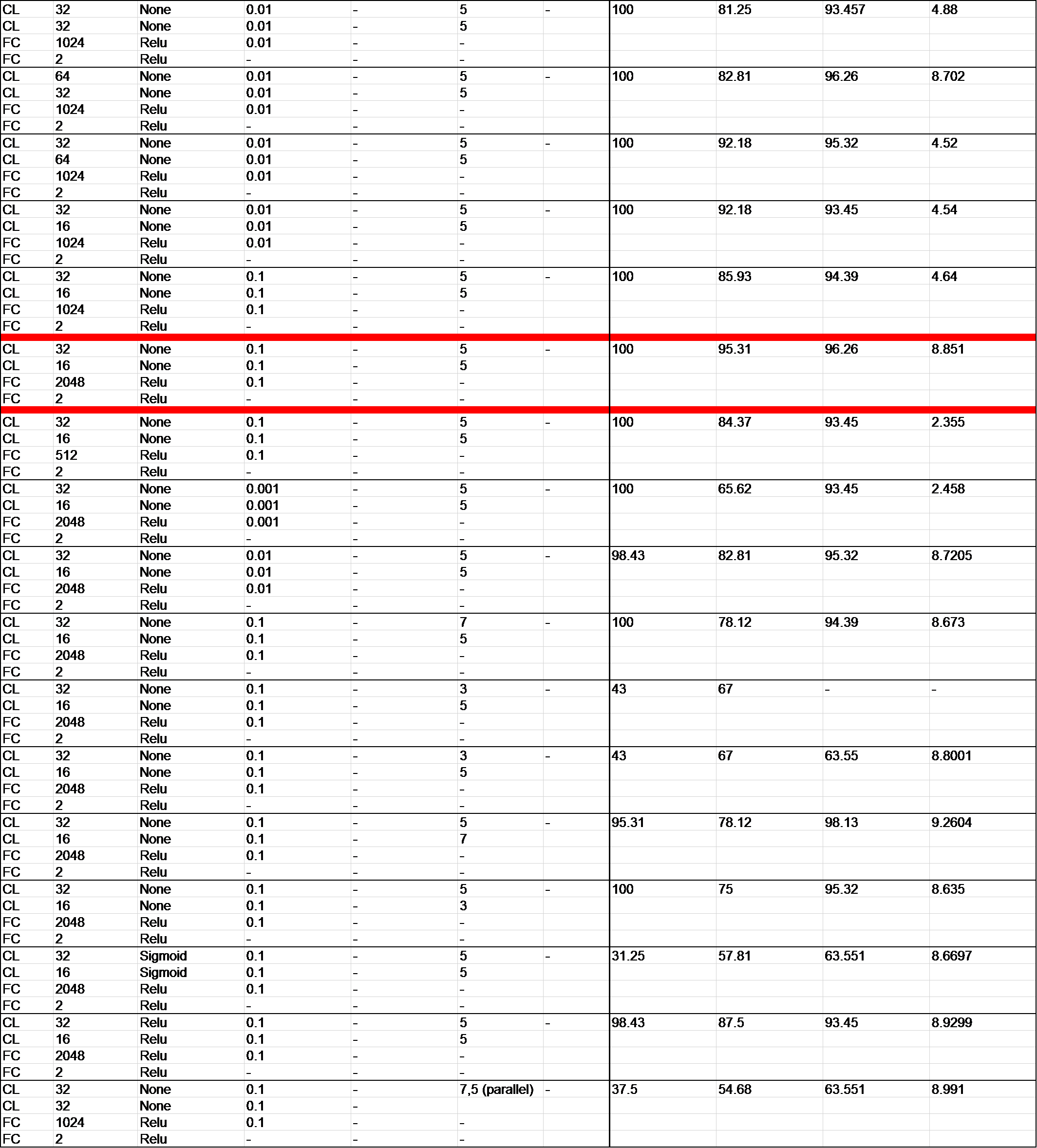} 
   \caption{Convolutional Neural Network Optimisation Process}
   \label{fig:OptimisationTable2}
\end{figure}

\begin{figure}
   \centering 
   \includegraphics[width=0.95\textwidth]{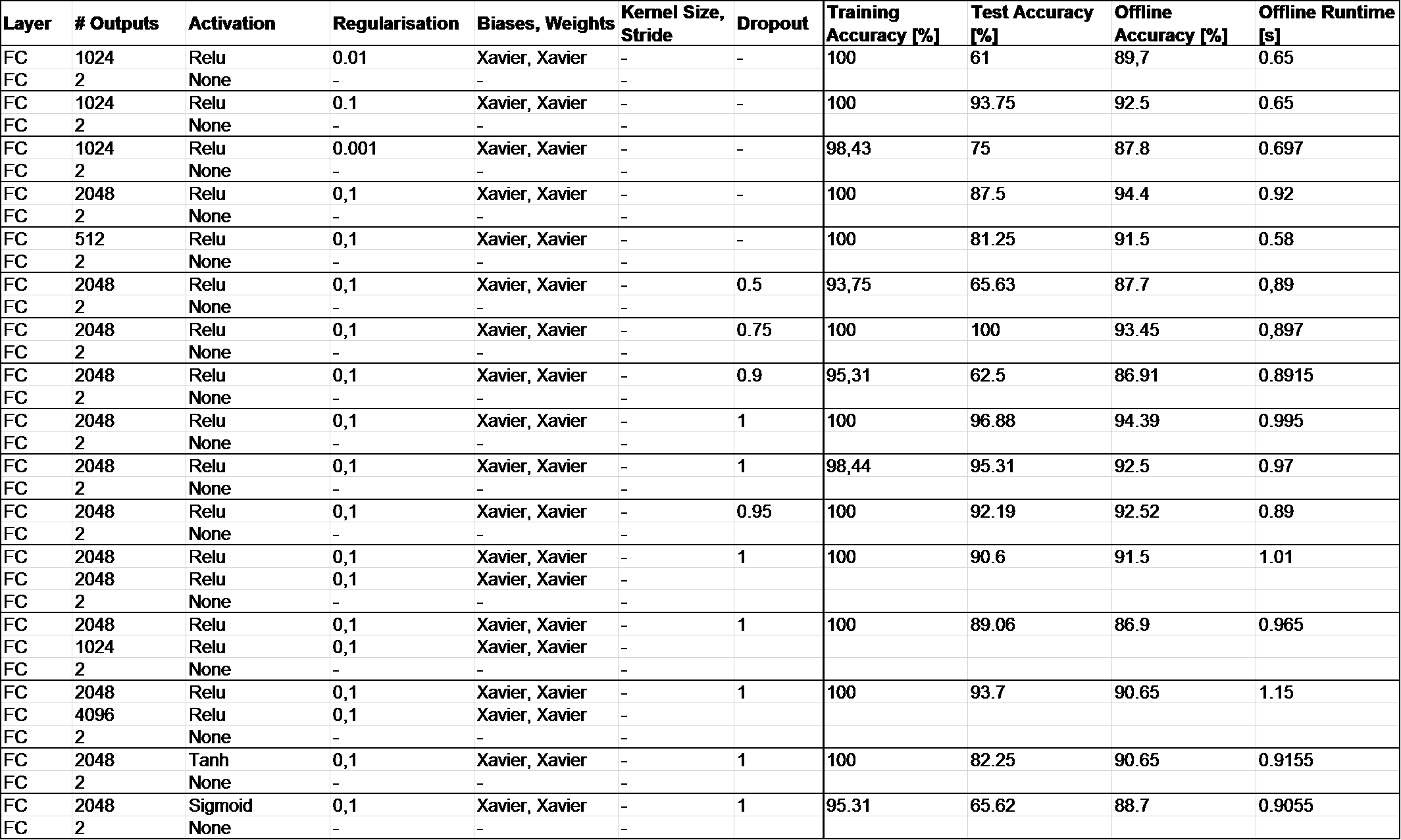} 
   \caption{Fully Connected Neural Network Optimisation Process}
   \label{fig:OptimisationTable1}
\end{figure}

Once the \ac{LiDAR} sensor has preselected the candidates and an image is cropped, the visual classification algorithm determines if the cropped image contains a cone or not. As stated in section \ref{sec:AlternativeApproaches}, we selected a neural network as a suitable visual classification approach for our application. 
\newline
Due to the distinguishable features of a cone and the necessity of the algorithm to run on a \ac{CPU} in real time, it was decided to evaluate two small neural networks; a fully connected neural network with a maximum of 3 layers and a small convolutional network. A cone with its orange colour and its shape can be easy differentiated from other objects so that a network containing merely a few layers is sufficient to detect cones. Furthermore, small networks require less time to classify an image and are therefore appropriate for running in real time on the computer on the vehicle. The fully connected network was selected due to its simplicity and the broad usage in different applications. As a fully connected network inputs the whole image at once, it is difficult to detect cones near to the camera as well as cones far away from the camera due the changing scale. In order to account for this problem, the convolutional network was selected as a second network for evaluation. The convolutional network extracts features at different levels and thereby is capable of detecting cones of different sizes in the cropped images. All the networks were programmed with the open-source software library tensorflow from Google \cite{tensorflow}.
\newline
When designing the two architectures, several optimisation iterations were performed. The parameters varied during the optimisation process can be found in Figure \ref{fig:OptimisationTable2} and Figure \ref{fig:OptimisationTable1}. The goal of the optimisation process was to increase accuracy and decrease the model's query time. As the final network should run in real time on the \ac{CPU} of the computer on the car, the computing time is an important criterion. 

\newpage

Different network architectures were trained on various changed parameters on a training dataset with over 20’000 images, which were recorded and labelled during this Bachelor Thesis. In total over 4 hours were spend on labelling data. The accuracy and the computing time of the neural networks was calculated and measured on a test as well as on a validation dataset with the same computer. In order to test the robustness of the neural networks, both datasets were recorded on a different day and under different light conditions than the training dataset. The information about the used datasets can be found in Tabel \ref{table:DatasetTable}.

\begin {table}[h]
	\begin{tabular}{|p{3.5cm}|p{3cm}|p{3cm}|p{3cm}|}
	\hline
	& Number of \newline Images & Number of \newline Positives & Number of \newline Negatives \\
	\hline
	Training Dataset & 20937 & 8249 & 12688 \\
	\hline
	Test Dataset & 710 & 429 & 281 \\
	\hline
	Validation Dataset & 108  & 68 & 40 \\
	\hline
	\end{tabular}
	\caption{Composition of the datasets}
	\label{table:DatasetTable}
\end {table}

\begin{figure}[h]
   \centering 
   \includegraphics[width=1.0\textwidth]{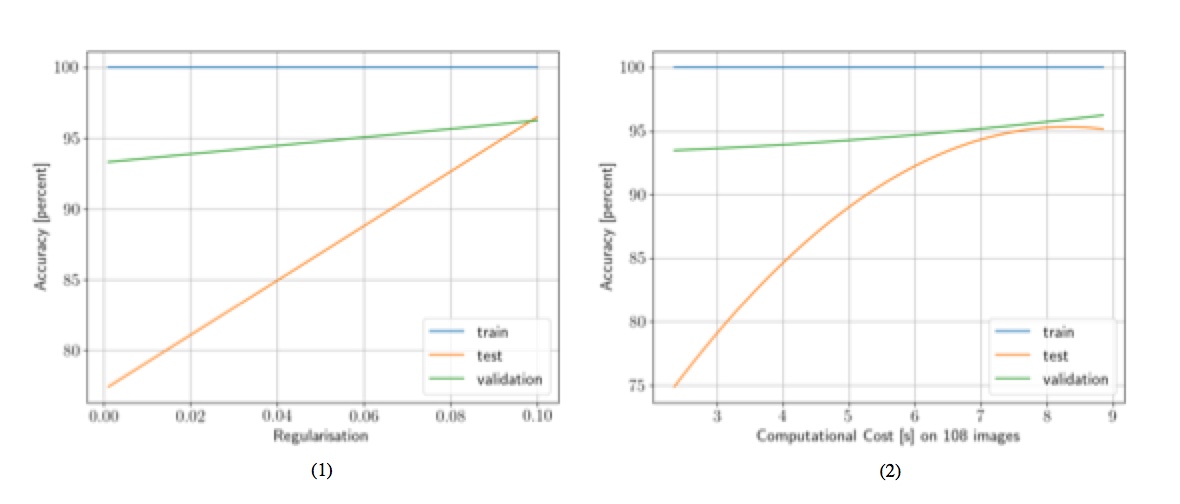} 
   \caption{Accuracy and regularisation factor graph (1), accuracy and computational time graph (2)}
   \label{fig:RegTimeGraphs}
\end{figure}

During the optimisation process, especially two adjustments had a great positive impact on the accuracy of all networks: the normalisation of the weights and transformation into the \ac{LAB} of the image colours as well as the introduction of regularisation functions. The cropped \ac{RGB} image is transformed into the \ac{LAB}, in which the colour information is separated from the brightness resulting in a more accurate classification of cones under different light conditions. As a second adjustment, in order to prevent the neural network from overfitting on the training dataset, the introduction of L1 regularisation functions has considerably improved the accuracy of the networks. This improvement can be seen in Figure \ref{fig:RegTimeGraphs} on the left side. Since the linear L1 regularisation reacts more robustly to outliers and as the network should recognise similiar objects, a L1 regularisation provides slightly more reliable results compared to L2 regularisation. With a higher regularisation factor, the accuracy on the test dataset increases. Moreover, a dropout was added, but did not improve the testing results. Furthermore, as illustrated in Figure \ref{fig:RegTimeGraphs} on the right side, the more time the evaluated neural network requires to classify images, the more accurate it is. From 9 s query time the model overfitting occurs, so that the accuracy on the validation dataset decreases. The computing time was measured by applying the neural network on 108 images on the validation dataset using the same computer for all datasets.

\begin{figure}[h]
   \centering 
   \includegraphics[width=0.8\textwidth]{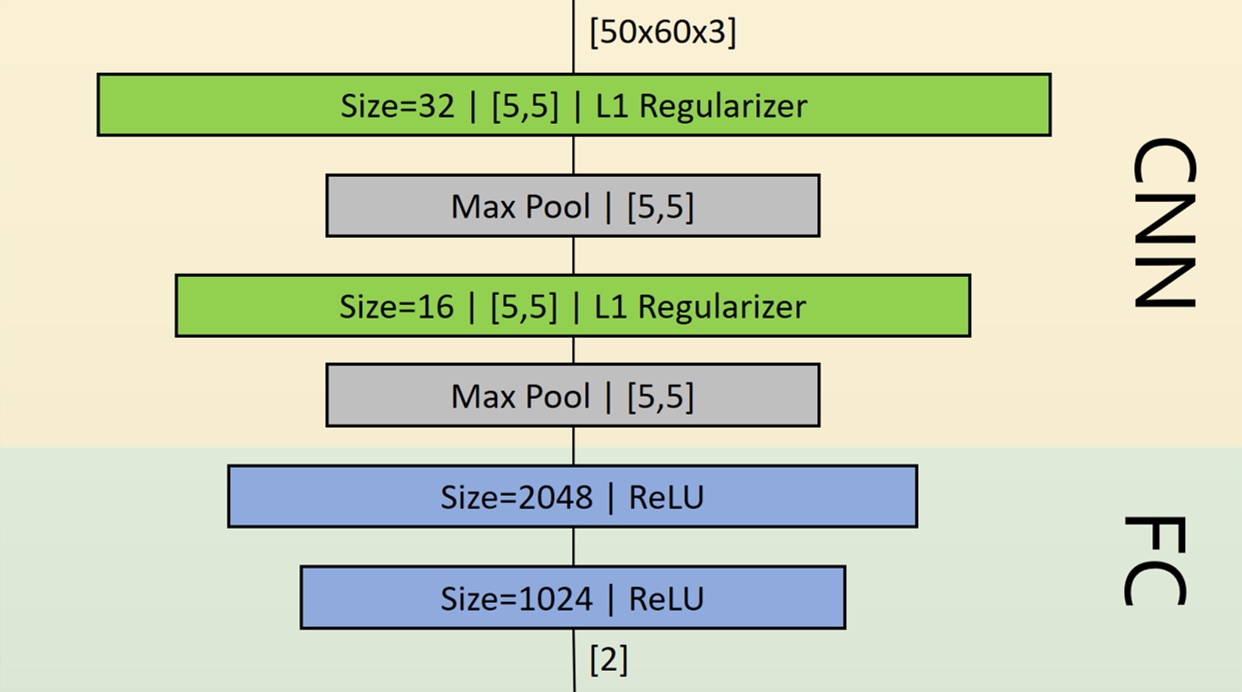} 
   \caption{Structure of the Convolutional Network}
   \label{fig:CCN}
\end{figure}

The final decision fell on a convolutional network, marked with red in Figure \ref{fig:OptimisationTable2}. This network had the highest accuracy on all datasets and with an average query time of 81 ms per image which is still capable of running in real time while driving. The final convolutional network was trained, as all the other evaluated neural networks, in 2000 iterations with a batch size of 64 and a learning rate of 0.001 on the training dataset. The structure of the final convolutional network is depicted in Figure \ref{fig:CCN}. 

\begin{figure}[h]
   \centering 
   \includegraphics[width=0.75\textwidth]{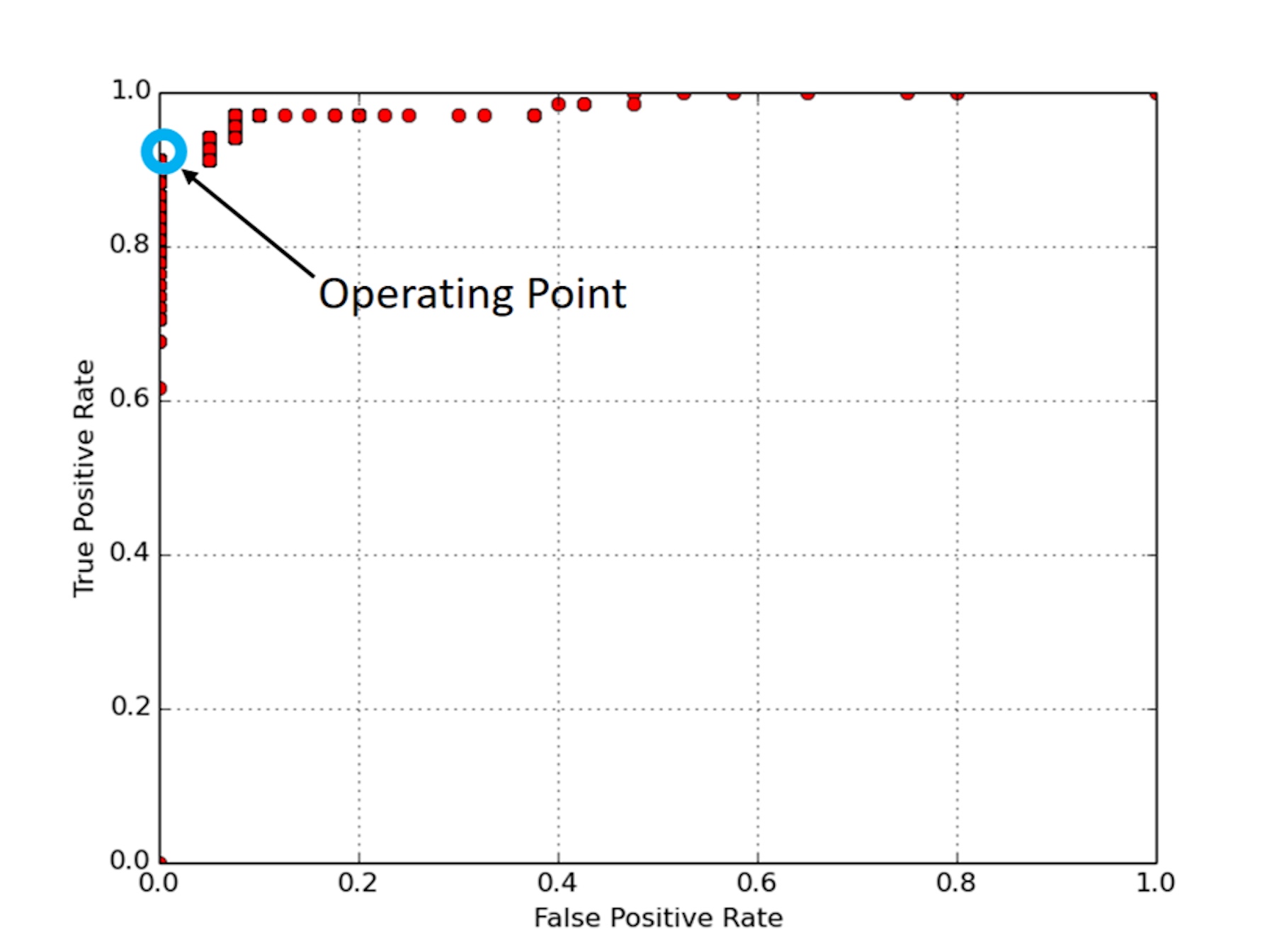} 
   \caption{ROC Curve}
   \label{fig:ROC}
\end{figure}

Moreover, in Figure \ref{fig:ROC} a \ac{ROC} curve is illustrated, which was created on the validation dataset. The operating point of the final convolutional network is marked with a blue circle.  The operating point was chosen in order to have the highest true positive rate given a low false positive rate. Therefore, the algorithm detects with a low possibility an object falsely as a cone. This misclassification should be avoided, as the whole system is required to be restarted once an object is falsely classified as a cone. On the other hand, when an image of a cone is not classified as cone, the colour camera and the \ac{LiDAR} sensor detection provide more candidates for the neural network to detect the cone.

\subsection{Gridmap}
When an object is detected and classified as a cone, the position of the cone is stored in a global 2D cone gridmap. Next to visualisation issues, the 2D cone gridmap is required for planning a global path and therefore for driving autonomously around the cones. In order to calculate the global position of new detected cone, the pose of the car as well as the position of the new detected cone in the coordinate frame of the \ac{LiDAR} sensor are needed. The pose information of the car is determined by an external state estimation. Furthermore, the local position of the cone measured by the \ac{LiDAR} sensor is transmitted once the visual classification has identified a cone. The coordinate origin of the global frame is the start position of the car. Combining the pose of the car and the local position results in a global position of the cone in the global coordinate frame. Due to the high precision of the \ac{LiDAR} sensor, in our case, it is sufficient to rely on the first detected position of a cone and to drop further position measurements belonging to the same cone afterwards, thus a statistical position estimation approach of cones was not considered.

\subsection{Path Planner}

\begin{figure}[h]
   \centering 
   \includegraphics[width=0.8\textwidth]{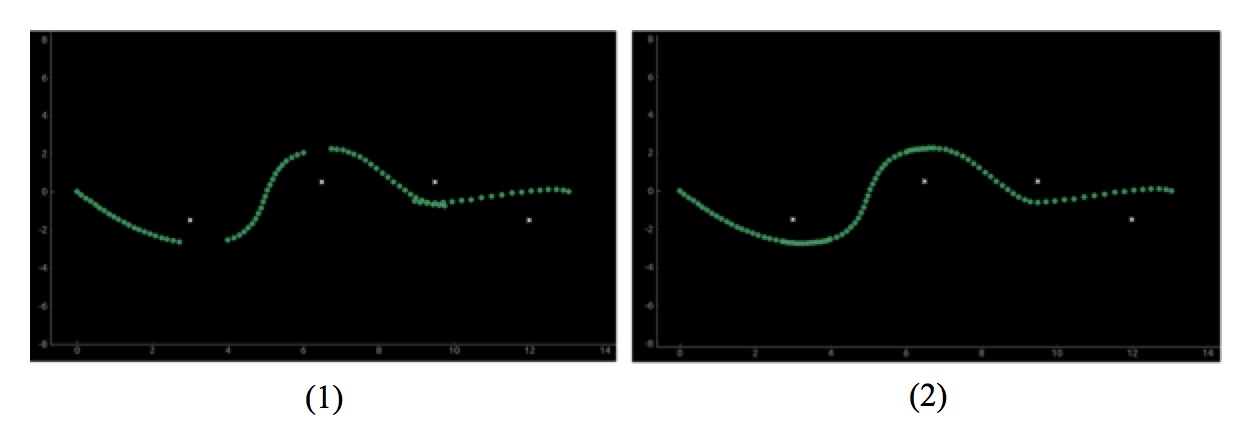} 
   \caption{Cosine Approximation and Path generated by the Path Planner}
   \label{fig:PathPlanner}
\end{figure}

In order to drive autonomously a slalom around the detected cones, the car has to follow a given path. This path is generated by the path planner. The path planner takes the 2D cone gridmap containing the detected cones as a single input and creates in the same global coordinate frame as the 2D cone gridmap a discrete path. The path has to be global due to the fact that the Pure Pursuit Controller adopted from Project \ac{ARC} needs a global and discrete path as well as a global position for calculating the steering and velocity controls. 
\newline
As the car should drive smoothly around the cones, the path needs to have a low second derivative. Furthermore, the path should pass the cone alternately on the right and left side of the driving direction. Due to these two requirements, a simple cosine approximation between two cones was chosen. As the \ac{LiDAR} sensor does not detect the cones in a specific order, it is therefore possible that a cone in nearer distance to the car is detected after the path is already generated around a previously detected cone further away. Due to this reason, the path planner also verifies if the newly detected cone lays between already detected cones. When this is the case, the path planner updates the whole path. 
\newline
As a first step, the path planner generates a straight forward path if no cones are detected. Once the first detected cone is stored in the 2D cone gridmap, the path planer generates a cosine in a range between 0 and $\pi$/4 rad from the start point to the first detected cone. Afterwards, the path planner rotates and stretches a cosine in a range of $\pi$ rad between two cones according to their global position. A simple cosine approximation is depicted on the left side of Figure \ref{fig:PathPlanner}. The green dots represent the calculated path and the white crosses symbolise the detected cones. When the cones are not aligned, two following cosines either have a gap or are overlapping, as shwon in \ref{fig:PathPlanner}. In the first case, the path planner adds filler points in the gap with a fixed distance to the selected cone. In the second case, the path planner removes the overlapping points. The filler points and the removal of points are illustrated in Figure \ref{fig:PathPlanner} on the right side.

\newpage
\section{Experimental Results}
\label{sec:PerformanceAnalysis}
The main focus of our Bachelor Thesis was to develop a reliable cone detection algorithm. As an additional objective, the car should drive fully autonomously around the detected cones. Therefore, the performance analysis can be divided into two sections, the evaluation of the cone detection and of the overall system performance.

\subsection{Cone Detection}
\begin{figure}[h]
   \centering 
   \includegraphics[width=0.9\textwidth]{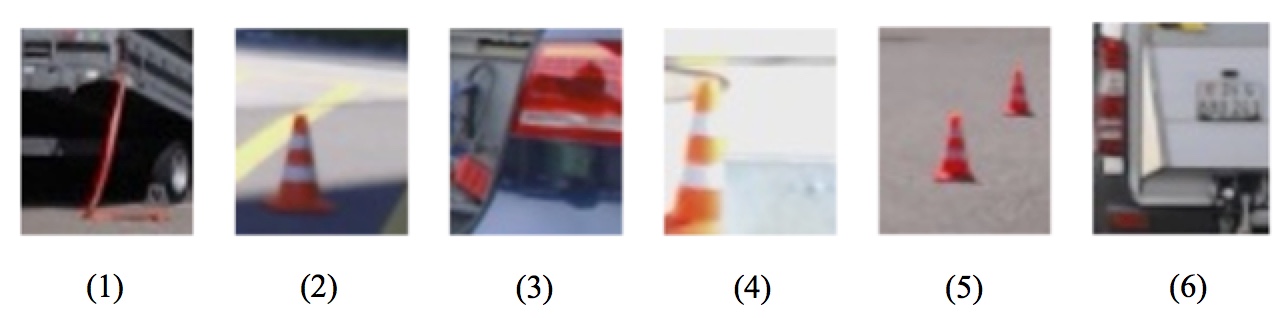} 
   \caption{Correct (1-5) and false (6) Classification}
   \label{fig:DataImages}
\end{figure}

The final convolutional network detects reliably cones amongst other objects. On the training dataset the accuracy reached 100\%, on the test dataset 95.3\% and on the validation dataset 96.26\%. In Figure \ref{fig:DataImages} on the left side, correctly classified example images can be seen. Only in very few specific cases, the convolutional network classifies an image falsely as a cone. One of these specific cases is depicted in Figure \ref{fig:DataImages} on the left side. The backlight of the car has the same red and white stripes as a cone. Furthermore, the shape of the backlight resembles the shape of a cone. 

\begin{figure}[h]
   \centering 
   \includegraphics[width=0.66\textwidth]{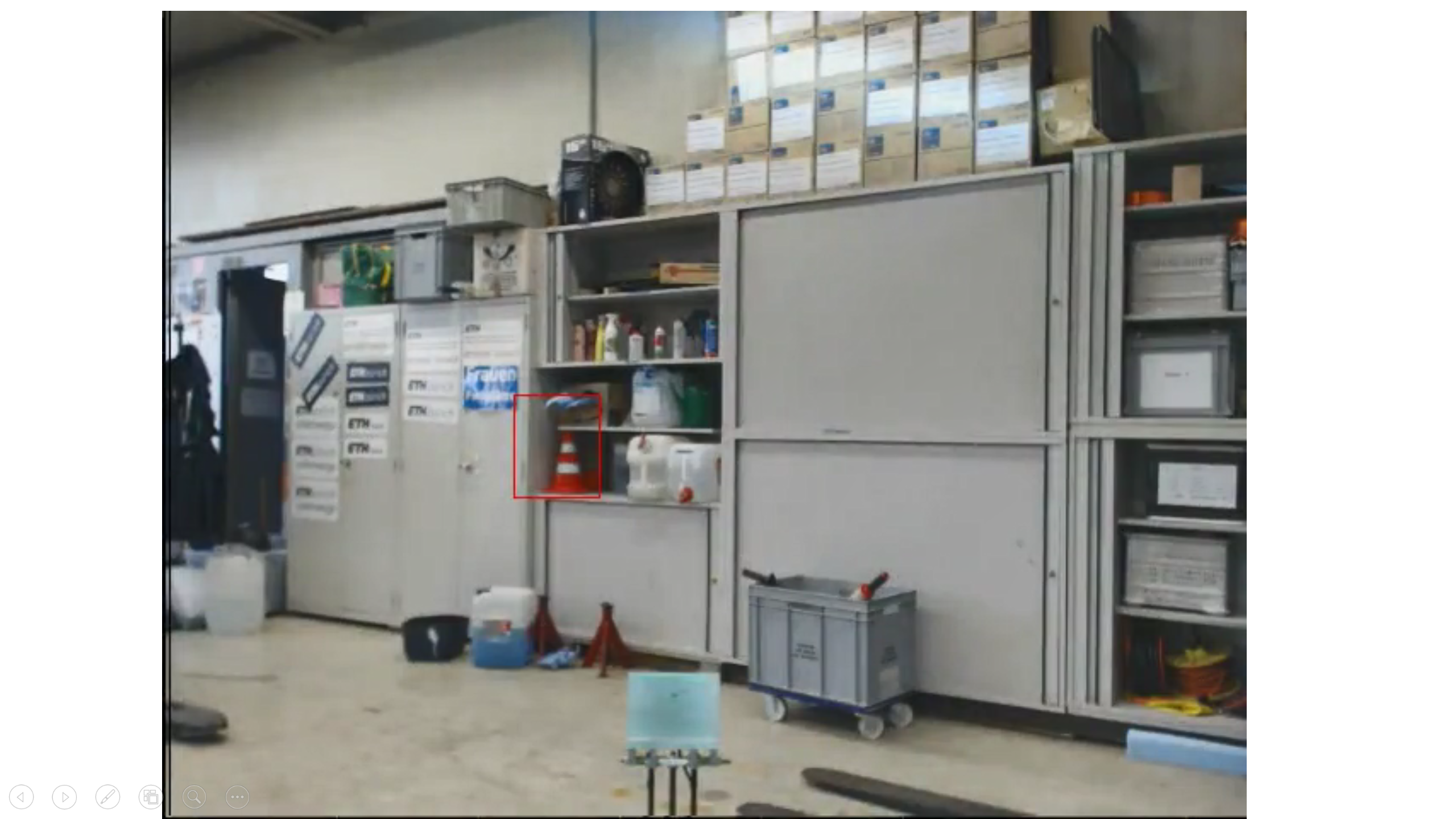} 
   \caption{Cone detected in a shelf with our cone detection algorithm}
   \label{fig:ConeShelves}
\end{figure}

As the final convolutional network is trained on images of cones on a road, a classification of cones in other surrounding would lead to a decreased accuracy. However, the network is still capable of detecting cones even in different surroundings than a road, which can be seen in Figure \ref{fig:ConeShelves}.  

Nevertheless, the final convolutional networks detects almost every cone placed on the road and with an average classification time of 81 ms per image, the convolutional network runs in real time on the computer on the car while driving. Moreover, the final convolutional network is capable of detecting cones even at velocities above 40 km/h. At higher velocities, the webcam captures only a few candidate images of the cones. The implemented networks still detects all the cones passed by the car. 
\newline
In comparison to the Inception v3 network (\cite{Inception}) \footnote{Inception v3 is deep neural network for image recognition, which is trained on ImageNet and can additionally be adapted to the recognition of specific objects (e.g. cones) \cite{InceptionWebsite}.}, which needs 371 ms to classify an image, the final convolutional network requires only 51 ms with the same computer. The value of the computing time of the final networks differs from the value stated in Figure \ref{fig:OptimisationTable2} due to the usage of two different computers. The accuracy of the Inception v3 network on the validation dataset could not be computed due to the required time to train the inception network on 20’000 images in order to compare both approaches properly. The performance compared to the methods mentioned in alternative Approaches can be found in Table \ref{table:Comparison}. In conclusion, the final convolutional network is capable of detecting reliably and rapidly the cones even when only few candidate images are provided.

\begin {table}[h]
	\begin{tabular}{|p{2.5cm}|p{2.5cm}|p{2.5cm}|p{2.5cm}|p{2.5cm}|}
	\hline
	Approaches & Accuracy [\%] & True Positive Rate [\%] & False Positive Rate [\%] & Time (per \newline image) [ms] \\
	\hline
	Colour & 84.3 & 89.7 & 25 & 0.07 \\
	\hline
	Angle & 63 & 100 & 100 & 1.73 \\
	\hline
	CNN & 95.4 & 94.1 & 2.5 & 50.99 \\
	\hline
	\end{tabular}
	\caption{Comparison alternative approaches}
	\label{table:Comparison}
\end {table}

As a further improvement for the cone detection, a filtering before the classification with the convolutional network using a combination of colour and line detection improves the computing time of the cone detection algorithm. One test with the combination was conducted and resulted in a lower computing time on the validation dataset. The results of the test are shown in Table \ref{table:Prefiltering}. The accuracy of the combined algorithm decreases as the line detection and colour detection are filtering some images with cones. As in our application the final neural network is sufficient regarding query time, we decided to maintain the merely neural network based classification approach.

\begin {table}[h]
	\begin{tabular}{|p{2.5cm}|p{2.5cm}|p{2.5cm}|p{2.5cm}|p{2.5cm}|}
	\hline
	Approaches & Accuracy [\%] & True Positive Rate [\%] & False Positive Rate [\%] & Time (per \newline image) [ms] \\
	\hline
	CNN & 95.4 & 94.1 & 2.5 & 50.99 \\
	\hline
	Filtered CNN & 91.7 & 88.2 & 2.5 & 36.54 \\
	\hline
	\end{tabular}
	\caption{Comparison CNN with filtered CNN}
	\label{table:Prefiltering}
\end {table}

\subsection{Overall System}

\begin{figure}[h]
   \centering 
   \includegraphics[width=0.9\textwidth]{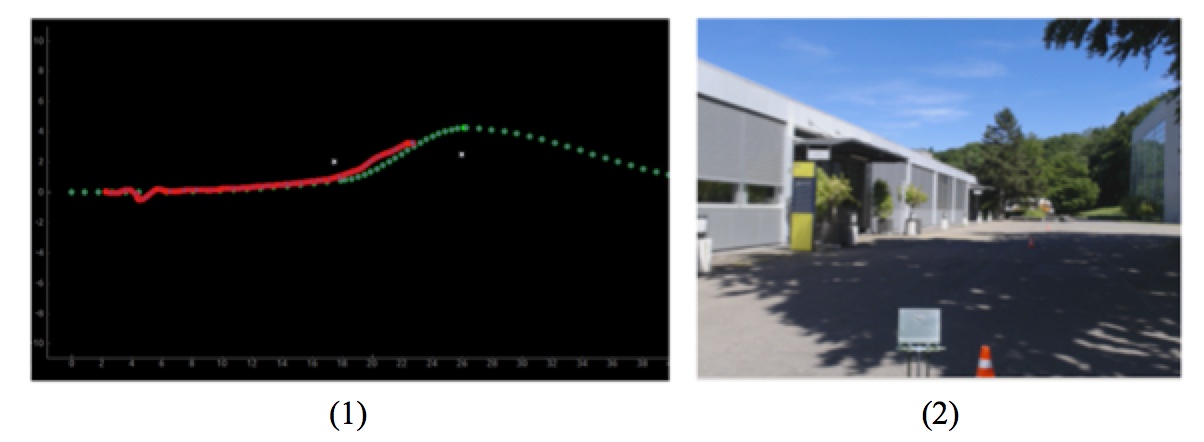} 
   \caption{Unsuccessful experiment using \ac{ROVIO} and \ac{ORB-SLAM2} }
   \label{fig:Comparison}
\end{figure}

The system developed within the scope of our Bachelor Thesis has the potential to drive fully autonomously around cones. Almost in every validation test at the end of the testing phase, the cone detection identified all the cones and the path planner generated a slalom around the cones accordingly. In one third of the tests, the car did drive successfully a slalom around two cones. The limiting factor was the external state estimation. At the beginning of the Bachelor Thesis we assumed, that the state estimation used during Project \ac{ARC} was accurate enough for the specific case of driving a slalom around cones. But as it turned out this assumption was not valid.
\newline
The visual odometry algorithm \ac{ROVIO}, optimised and used successfully during Project \ac{ARC}, had some global inaccuracies. As the drift is identical in teach as well as in repeat phase, during Project \ac{ARC}, the global drift does not pose a problem. Due to the global inaccuracy, \ac{ROVIO} was not suitable for driving a narrow path around cones. One of the failed experiment with \ac{ROVIO} in combination with the visual \ac{SLAM} algorithm \ac{ORB-SLAM2} can be seen in Figure \ref{fig:Comparison}. On the left side, the \ac{GUI} is depicted with the generated path marked with green dots and the driven path with red dots. The position of the detected cones are illustrated with white crosses. On the \ac{GUI}, the position of the car and the cones seems to be correct. However, as it can be seen on the colour camera image of the same test on the right side of Figure \ref{fig:Comparison}, the car would have driven over the cone. 

The localisation with the \ac{GNSS}-Receiver was tested as well but lead to an imprecise position estimation. We thereby decided to use a simple kinematic bicycle vehicle model. This method takes only the actual steering angle and the velocity measured by the rotary encoders as inputs. Unfortunately, due to the inaccuracies of the kinematic bicycle vehicle model, the accuracy of the position estimation decreases while driving. Therefore, we were only able to drive a short distance. In retro perspective, an earlier testing of the state estimation would have shown the problem at an earlier stage and would resulted in more time in order to optimise \ac{ROVIO} for our specific case. However, the system was capable of driving around two cones using the kinematic bicycle vehicle model in certain cases. In conclusion, the development of a reliable cone detection as well as of the framework for driving a slalom around cones could be achieved in this Bachelor Thesis.

\newpage
\section{Conclusion}
\label{sec:Conclusion}

\subsection{Summary}

In this Bachelor Thesis an innovative approach for performing object detection was demonstrated using a combination of a mono colour camera and a \ac{LiDAR} sensor, illustrated by detecting cones. This approach allows to combine a reliable (i.e. as shown computational demanding) classification with an accurate position estimation of the detected object and is capable to run in a real-time application. For cone detection, easy contraints for each laser point to preselect cone candidates and a convolutional neural network to classify the candidates provide an optimal result. In this way, a high recall as well as a high accuracy in object detection were achieved.

\subsection{Outlook}

Regarding the overall system, the car drove a short distance around cones due to the inaccurate state estimation. In the state estimation used for our Bachelor Thesis we see a lot of room for improvement as the state estimation was not in the focus of this Bachelor Thesis and assumed to be given. An optimisation of \ac{ROVIO} for our specific case would probably increase the accuracy of the pose estimation. It is also possible to localise against the computed position of already detected cones. In combination with \ac{ROVIO}, a localisation using the already detected cones would lead to a more robust state estimation. Furthermore, a \ac{MPC} would improve path planning as it would account for the dynamics of the car. The car would then drive on an optimised path around cones. Moreover, it is worth considering to position the \ac{LiDAR} sensor at the front of the car in order to detect the cones constantly without having a blind spot. As a result not only a more robust state estimation but also the possibility of a local control would emerge. We are confident, that with these mentioned modifications and our reliable cone detection a system can be developed capable of driving a long distance around cones even at higher velocities.

\newpage
\bibliographystyle{bibliography/IEEEtranN}
\bibliography{bibliography/references}
\addcontentsline{toc}{section}{Bibliography}
\newpage
\listoffigures
\listoftables
\addcontentsline{toc}{section}{List of Figures \& Tables}

\end{document}